\documentclass[onecolumn,peerreview,11pt]{IEEEtran}
\usepackage{amsmath,epsfig,amsfonts,amssymb}
\usepackage{booktabs}
\usepackage{algorithm}
\usepackage{algorithmic}

\usepackage{pgf}
\usepackage{tikz}
\usepackage{pgfplots}
\usepackage{setspace}
\usepackage{hyperref}
\usepgfplotslibrary{ternary} 
\newcommand{\algorithmicbreak}{\textbf{break}}
\newcommand{\BREAK}{\STATE \algorithmicbreak}

\newcommand{\rev}[1]{{\em \color{red} #1}}

\newtheorem{prop}{Proposition}

\begin{document}
\title{Fast forward feature selection for the nonlinear classification
  of hyperspectral images}

\author{Mathieu   Fauvel,   Clement   Dechesne,  Anthony   Zullo   and
  Fr{\'e}d{\'e}ric Ferraty%
  \thanks{M.   Fauvel,  C.   Deschene  and  A.   Zullo  are  with  the
    Universit{\'e} de  Toulouse, INP-ENSAT,  UMR 1201  DYNAFOR, France
    and with the INRA, UMR 1201 DYNAFOR, France}%
  \thanks{A.    Zullo   and  F.    Ferraty   are   with  Institut   de
    Math{\'e}matiques de Toulouse - UMR  5219 IMT \& Universit{\'e} de
    Toulouse, France}%
  \thanks{This  work was  supported  by the  French National  Research
    Agency  (ANR)  under  Project Grant  ANR-13-JS02-0005-01  (Asterix
    project), by the EFPA department  of the French National Institute
    for  Agricultural Research  (INRA)  trough  an Innovative  Project
    grant,  by  the French  National  Spacial  Agency (CNES)  and  the
    Midi-Pyr{\'e}n{\'e}es region.}}
\maketitle
\begin{abstract}
  A  fast forward  feature selection  algorithm is  presented in  this
  paper.  It  is based on  a Gaussian mixture model  (GMM) classifier.
  GMM are  used for  classifying hyperspectral images.   The algorithm
  selects   iteratively   spectral    features   that   maximizes   an
  \rev{estimation of  the classification rate. The  estimation is done
    using the k-fold  cross validation}.  In order to  perform fast in
  terms   of   computing   time,  an   efficient   implementation   is
  proposed. First, the  GMM can be updated when the  estimation of the
  classification rate  is computed,  rather than re-estimate  the full
  model.  Secondly, using  marginalization of the GMM,  sub models can
  be  directly obtained  from  the  full model  learned  with all  the
  spectral   features.   Experimental   results  for   two  \rev{real}
  hyperspectral data sets  show that the method performs  very well in
  terms of classification accuracy  and processing time.  Furthermore,
  the extracted model contains very few spectral channels.
\end{abstract}

\IEEEpeerreviewmaketitle
\section{Introduction}
\label{sec:intro}

Since the pioneer paper of J. Jimenez and D.  Landgrebe~\cite{661089},
it is  well known that  hyperspectral images need  specific processing
techniques      because      conventional      ones      made      for
multispectral/panchromatic images  do not adapt  well to hyperspectral
images. Generally speaking, the increasing number of spectral channels
poses   theoretical    and   practical   problems~\cite{donoho}.    In
particular,  for the  purpose  of pixel  classification, the  spectral
dimension  needs  to  be  handle  carefully because  of  the  ``Hughes
phenomenon''~\cite{hughes}:  with  a limited  training  set, beyond  a
certain  number of  spectral features,  a reliable  estimation  of the
model parameters is not possible.

Many works  have been published  \rev{since the 2000s} to  address the
problem  of classifying  hyperspectral images.  A non-exhaustive  list
should include  techniques from  the machine learning  theory (Support
Vector         Machines,         Random         Forest,         neural
networks)~\cite{fauvel:hal-00737075}, statistical models~\cite{661089}
and  dimension   reduction~\cite{DR:guided:tour}.   SVM,   and  kernel
methods   in   general,   have  shown   remarkable   performances   on
hyperspectral      data      in      terms      of      classification
accuracy~\cite{kernel:methods:rs}.  However, these  methods may suffer
from a high computational load and  the interpretation of the model is
usually not trivial.

In parallel to  the emergence of kernel methods, the  reduction of the
spectral dimension has received a  lot of attention.  According to the
absence or  presence of training  set, the dimension reduction  can be
unsupervised or supervised.  The former  try to describe the data with
a  lower  number of  features  that  minimize a  reconstruction  error
measure, while  the latter try  to extract features that  maximize the
separability  of  the classes.   One  of  the most  used  unsupervised
feature  extraction   method  is  the  principal   component  analysis
(PCA)~\cite{661089}.  But  it has  been demonstrated  that PCA  is not
optimal for the  purpose of classification~\cite{1294808}.  Supervised
method, such as  the Fisher discriminant analysis  or the non-weighted
feature extraction  have shown  to perform better  for the  purpose of
classification.    Other  feature   extraction  techniques,   such  as
independent  component  analysis~\cite{5942156},   have  been  applied
successfully and demonstrate  that even SVM can  benefits from feature
reduction~\cite{fauvel:hal-00283769,6851112}.   However,  conventional
supervised techniques suffer from similar problems than classification
algorithms in high dimensional space.

Rather  than  supervised and  unsupervised  techniques,  one can  also
distinguish   dimension   reduction  techniques   into   \emph{feature
  extraction}  and   \emph{feature  selection}.    Feature  extraction
returns a linear/nonlinear combination of the original features, while
feature selection  returns a  subset of  the original  features. \rev{
  While  feature  extraction and  feature  selection  both reduce  the
  dimensionality of  the data, the  latter is much  more interpretable
  for the  end-user}.  The  extracted subset  corresponds to  the most
important features  for the  classification, i.e., the  most important
wavelengths. \rev{For  some applications, these spectral  channels can
  be     used     to      infer     mineralogical     and     chemical
  properties~\cite{6947182}.}

Feature  selection   techniques  generally  need  a   criterion,  that
evaluates  how  the  model  built  with a  given  subset  of  features
performs, and an optimization procedure  that tries to find the subset
of  features  that maximizes/minimizes  the  criterion~\cite{4069122}.
Several methods  have been  proposed according  to that  setting.  For
instance,  an  entropy  measure  and a  genetic  algorithm  have  been
proposed  in~\cite[Chapter  9]{chein2007hyperspectral}, but  the  band
selection  was  done  independently   of  the  classifier,  i.e.,  the
criterion  was not  directly related  to the  classification accuracy.
\rev{Jeffries   Matusita  (JM)   distance  and   steepest-ascent  like
  algorithms were proposed in~\cite{934069}.  The method starts with a
  conventional  sequential  forward   selection  algorithm,  then  the
  obtained set of features is  updated using local search.  The method
  has been extended  to take into account  spatial variability between
  features in~\cite{5161332} where a multiobjective criterion was used
  to  take  into  account  the  class  separability  and  the  spatial
  variability.   JM distance  and exhaustive  search as  well as  some
  refinement  techniques have  been proposed  also in~\cite{4069122}.}
However  rather  than  extracting  spectral  features,  the  algorithm
returns the average  over a certain bandwidth  of contiguous channels,
which can make the interpretation  difficult and often leads to select
a large part of the electromagnetic spectrum. \rev{Similarly, spectral
  intervals  selection  was   proposed  in~\cite{4317535},  where  the
  criterion  used was  the square  representation error  (square error
  between the  approximate spectra and  the original spectra)  and the
  optimization problem was solved using dynamic programming. These two
  methods reduce the dimensionality of the data, but cannot be used to
  extract  spectral  variables.}    Recently,  forward  selection  and
genetic algorithm driven by  the classification error minimization has
been proposed in~\cite{lebris:fs}.

\rev{Feature selection  has been also  proposed for kernel  methods. A
  recursive scheme used to remove  features that exhibit few influence
  on  the  decision   function  of  a  nonlinear   SVM  was  discussed
  in~\cite{tuia2009classification}.  Alternatively, a shrinkage method
  based on $\ell_1$-norm and linear  SVM has been investigated by Tuia
  \emph{et  al.}~\cite{tuia2014automatic}.   The  authors  proposed  a
  method where the features are extracted during the training process.
  However,  to make  the method  tractable in  terms of  computational
  load, a linear model is used  for the classification, that can limit
  the discriminating  power of  the classifier.   In~\cite{5440922}, a
  dependence measure between spectral features and thematic classes is
  proposed using kernel evaluation.  The  measure has the advantage to
  apply  to  multiclass  problem  making  the  interpretation  of  the
  extracted features easier.}

Feature  selection   usually  provides   good  results  in   terms  of
classification accuracy.  However, several drawbacks can be identified
from the above mentioned literature:
\begin{itemize}
\item  It can  be very  time consuming,  in particular  when nonlinear
  classification models are used.
\item  When linear  models are  used  for the  selection of  features,
  performances in terms of  classification accuracy are not satisfying
  and therefore another nonlinear classifier  should be used after the
  feature extraction.
\item  \rev{For  multiclass  problem,  it is  sometimes  difficult  to
    interpret  the  extracted features  when  a  collection of  binary
    classifiers is used (e.g., SVM).}
\end{itemize}

In  this  work, it  is  proposed  to  use  a forward  strategy,  based
on~\cite{Ferraty01122010},  that  uses   an  efficient  implementation
scheme and allows to process a large  amount of data, both in terms of
number of  samples and  variables. The method,  called \emph{nonlinear
  parsimonious  feature  selection}   (NPFS),  selects  iteratively  a
spectral feature  from the original set  of features and adds  it to a
pool of  selected features.   This pool  is used  to learn  a Gaussian
mixture  model (GMM)  and  each  feature is  selected  according to  a
classification rate. The  iteration stops when the  increased in terms
of classification rate is lower than  a user defined threshold or when
the maximum  number of  features is reached.   In comparison  to other
feature extraction algorithms,  the main contributions of  NPFS is the
ability to select spectral features through a nonlinear classification
model and its high computational efficiency. Furthermore, NPFS usually
extracts  a very  few  number of  features  (lower than  5  \% of  the
original number of spectral features).

The   remaining    of   the    paper   is   organized    as   follows.
Section~\ref{sec:model}  presents  the  algorithm  with  the  Gaussian
mixture model  and the efficient  implementation. Experimental results
on  three  hyperspectral data  sets  are  presented  and discussed  in
Section~\ref{sec:exp}.   Conclusions  and  perspectives  conclude  the
paper in Section~\ref{sec:conclusion}.

\section{Non linear parsimonious feature selection}
\label{sec:model}
The  following notations  are  used  in the  remaining  of the  paper.
$\mathcal{S}=\big\{\mathbf{x}_i,y_i\big\}_{i=1}^n$ denotes  the set of
training   pixels,   where   $\mathbf{x}_i\in   \mathbb{R}^d$   is   a
$d$-dimensional  pixel vector,  $y_i=1,\ldots,C$ is  its corresponding
class, $C$  the number  of classes, $n$  the total number  of training
pixels and $n_c$ the number of training pixels in class $c$.
\subsection{Gaussian mixture model}
For a Gaussian  mixture model, it is supposed  that the observed pixel
is a realization of a $d$-dimensional random vector such as
\begin{eqnarray}
  p(\mathbf{x})=\sum_{c=1}^C\pi_cp(\mathbf{x}|c),
\end{eqnarray}
where $\pi_c$ is the proportion  of class $c$ ($0\leq \pi_c\leq 1$ and
$\sum_{c=1}^C\pi_c=1$)  and  $p(\mathbf{x}|c)$  is  a  $d$-dimensional
Gaussian distribution, i.e.,
$$p(\mathbf{x}|c)=\frac{1}{(2\pi)^{d/2}|\boldsymbol{\Sigma}_c|^{1/2}}\exp\left(-\frac{1}{2}(\mathbf{x}-\boldsymbol{\mu}_c)^T\boldsymbol{\Sigma}_c^{-1}(\mathbf{x}-\boldsymbol{\mu}_c)\right).$$
with  $\boldsymbol{\mu}_c$  being  the   mean  vector  of  class  $c$,
$\boldsymbol{\Sigma}_c$ being  the covariance matrix of  class $c$ and
$|\boldsymbol{\Sigma}_c|$  its  determinant.   Following  the  maximum
\emph{a posteriori} rule, a given pixel is classified to the class $c$
if  $p(c|\mathbf{x})\geq  p(k|\mathbf{x})$   for  all  $k=1,\ldots,C$.
Using the Bayes formula, the posterior probability can be written as
\begin{eqnarray}
  p(c|\mathbf{x})=\frac{\pi_cp(\mathbf{x}|c)}{\sum_{k=1}^C\pi_kp(\mathbf{x}|k)}.
\end{eqnarray}
Therefore, the  maximum  \emph{a  posteriori} rule can be written as 
\begin{eqnarray}\label{posterior}
  \mathbf{x} \text{ belongs to }c \Leftrightarrow c= \arg\max_{k=1,\ldots,C}\pi_kp(\mathbf{x}|k).
\end{eqnarray}
By taking  the log of  eq.~(\ref{posterior}) the final decision  rule is
obtained (also known as quadratic discriminant function)
\begin{eqnarray}\label{eq:predict}
  Q_c(\mathbf{x})= -(\mathbf{x}-\boldsymbol{\mu}_c)^T\boldsymbol{\Sigma}_c^{-1}(\mathbf{x}-\boldsymbol{\mu}_c) -\ln(|\boldsymbol{\Sigma}_c|)+2\ln(\pi_c).
\end{eqnarray}
Using standard  maximization of  the log-likelihood, the  estimator of
the model parameters are given by
\begin{eqnarray}
  \hat{\pi}_c&=&\frac{n_c}{n},\label{eq:prop}\\
  \hat{\boldsymbol{\mu}}_c&=&\frac{1}{n_c}\sum_{i=1}^{n_c}\mathbf{x}_i,\label{eq:mean}\\
  \hat{\boldsymbol{\Sigma}}_c&=&\frac{1}{n_c}\sum_{i=1}^{n_c}(\mathbf{x}_i-\hat{\boldsymbol{\mu}}_c)(\mathbf{x}_i-\hat{\boldsymbol{\mu}}_c)^T.\label{eq:cov}
\end{eqnarray}
with $n_c$ is the number of sample of class $c$.

For GMM, the ``Hughes phenomenon'' is related to the estimation of the
covariance matrix. If the number of training samples is not sufficient
for  a good  estimation  the computation  of the  inverse  and of  the
determinant  in eq.(\ref{eq:predict})  will be  very \rev{numerically}
unstable, leading  to poor  classification accuracy. For  instance for
the covariance matrix,  the number of parameters to  estimate is equal
to $d(d+1)/2$: if $d=100$ then  $5050$ parameters have to be estimated
then the minimum  number of training samples for  the considered class
should be  at least $5050$. Note  in that case the  estimation will be
possible but not accurate. \rev{Feature selection tackles this problem
  by allowing  the construction of  GMM with  a reduced number  $p$ of
  variables, with $p<<d$ and $p(p+1)/2<n_c$.}

\subsection{Forward feature selection}
The   forward   feature   selection  works   as   follow~\cite[Chapter
3]{hastie2001elements}.   It starts  with  an empty  pool of  selected
features. At each  step, the feature that most  improves an estimation
of the classification rate is added  to the pool.  The algorithm stops
either if the increase of the estimated  classification rate is
too low or if the maximum number of features is reached.

The  $k$-fold  cross-validation  ($k$-CV)  is used  in  this  work  to
estimate the classification rate.  To  compute the $k$-CV, a subset is
removed from $\mathcal{S}$  and the GMM is learned  with the remaining
training samples.  A test error  is computed with the removed training
samples used as validation samples.  The process is iterated $k$ times
and the  estimated classification  rate is computed  as the  mean test
error over the $k$ subsets of $\mathcal{S}$.

The efficient implementation  of the NPFS relies on  a fast estimation
of  the parameters  of the  GMM when  the $k$-CV  is computed.  In the
following,  it  will be  shown  that  by  using  update rules  of  the
parameters  and   the  marginalization  properties  of   the  Gaussian
distribution, it is  possible to perform $k$-CV  and forward selection
quickly.  As  a consequence, the  GMM model  is learned only  one time
during  the   whole  training  step.    The  algorithm~\ref{algo:npfs}
presents a pseudo code of the proposed method.

\subsubsection{Fast estimation of the model on $\mathcal{S}^{n-\nu}$}
In this  subsection, it  is shown  that each  parameter can  be easily
updated when a subset is taken off $\mathcal{S}$.  
\begin{prop}[Proportion]
  The update rule for the proportion is
  \begin{eqnarray}\label{update:prop}\hat{\pi}_c^{n-\nu}=\frac{n\hat{\pi}_c^n-\nu_c}{n-\nu}\end{eqnarray}
  where $\hat{\pi}_c^{n-\nu}$ and  $\hat{\pi}_c^n$ are the proportions
  of class  $c$ computed over  $n-\nu$ and $n$ respectively,  $\nu$ is
  the number  of removed  samples from  $\mathcal{S}$, $\nu_c$  is the
  number    of   removed    samples   from    class   $c$    such   as
  $\sum_{c=1}^C\nu_c=\nu$.
\end{prop}
\begin{prop}[Mean vector]
  The update rule for the mean vector is
  \begin{eqnarray}\label{update:mean}\hat{\boldsymbol{\mu}}_c^{n_c-\nu_c}                     =
    \frac{n_c\hat{\boldsymbol{\mu}}_c^{n_c}-\nu_c\hat{\boldsymbol{\mu}}_c^{\nu_c}}{n_c-\nu_c}\end{eqnarray}
  where $\hat{\boldsymbol{\mu}}_c^{n_c}$ and $\hat{\boldsymbol{\mu}}_c^{n_c-\nu_c}$ are the mean vectors of class  $c$ computed over the $n_c$ and  $n_c-\nu_c$ training samples respectively, $\hat{\boldsymbol{\mu}}_c^{\nu_c}$ is the mean vector of the $\nu_c$ removed samples from class $c$.
\end{prop}

\begin{prop}[Covariance matrix]
  The update rule for the covariance matrix is 
\begin{eqnarray}\label{update:cov}
  \hat{\boldsymbol{\Sigma}}^{n_c-\nu_c}_c = \frac{n_c}{(n_c-\nu_c)}\hat{\boldsymbol{\Sigma}}^{n_c}_c-\frac{\nu_c}{(n_c-\nu_c)}\hat{\boldsymbol{\Sigma}}^{\nu_c}_c - \frac{n_c\nu_c}{(n_c-\nu_c)^2}\left(\hat{\boldsymbol{\mu}}_c^{n_c}-\hat{\boldsymbol{\mu}}_c^{\nu_c}\right)\left(\hat{\boldsymbol{\mu}}_c^{n_c}-\hat{\boldsymbol{\mu}}_c^{\nu_c}\right)^T
\end{eqnarray}
where                 $\hat{\boldsymbol{\Sigma}}^{n_c}_c$                and
$\hat{\boldsymbol{\Sigma}}^{n_c-\nu_c}_c$  are  the covariance  matrices  of
class $c$  computed over  the $n_c$  and $n_c-\nu_c$  training samples
respectively.
\end{prop}

\subsubsection{Particular case of leave-one-out cross-validation}
When  very  few  training  samples  are  available,  it  is  sometimes
necessary to resort to leave-one-out cross-validation ($k=n$). Updates
rules are still valid, but it is also possible to get a fast update of
the decision function. If the removed  sample does not belong to class
$c$,  only  the  proportion  term  in  eq.~(\ref{eq:predict})  change,
therefore the updated decision rule can be written as:
\begin{eqnarray}\label{update:decision}
  Q_c^{n_c-1}(\mathbf{x}_n)=Q_c^{n_c}(\mathbf{x}_n) +2\ln\big(\frac{n-1}{n}\big).
\end{eqnarray}
\rev{where $Q_c^{n_c}$  and $Q_c^{n_c-1}$  are the decision  rules for
  class $c$ computed  with $n_c$ and $n_c-1$  samples respectively. If
  the removed sample $\mathbf{x}_n$ belongs  to class $c$ then updates
  rules become:
\begin{prop}[Proportion-loocv]
  \begin{eqnarray}\hat{\pi}_c^{n-1}=\frac{n\hat{\pi}_c^n-1}{n-1}\end{eqnarray}
\end{prop}
\begin{prop}[Mean vector-loocv]
  \begin{eqnarray}
    \hat{\boldsymbol{\mu}}_c^{n_c-1} = \frac{n_c\hat{\boldsymbol{\mu}}_c^{n_c}-\mathbf{x}_{n}}{n_c-1}
  \end{eqnarray}
\end{prop}
\begin{prop}[Covariance matrix-loocv]
  \begin{eqnarray}
  \begin{array}{rcl}
    \hat{\boldsymbol{\Sigma}}_c^{{n_c}-1} &= &\dfrac{{n_c}}{{n_c}-1}\hat{\boldsymbol{\Sigma}}_c^{{n_c}}- \dfrac{{n_c}}{({n_c}-1)^2}\left(\mathbf{x}_{{n}} -\hat{\boldsymbol{\mu}}_c^{n_c}\right)\left(\mathbf{x}_{{n}} -\hat{\boldsymbol{\mu}}_c^{n_c}\right)^T.
  \end{array}
\end{eqnarray}
\end{prop}
where $n_c-1$  denotes that the  estimation is done with  only $n_c-1$
samples rather  than the  $n_c$ samples of  the class.}

An update rule for the case where the sample belongs the class $c$ can
be  written by  using  the Cholesky  decomposition  of the  covariance
matrix and rank-one downdate, but the downdate step is not numerically
stable and not used here.

\subsubsection{Marginalization of Gaussian distribution}
To get the GMM model over a subset of the original set of features, it
is  only necessary  to drop  the non-selected  features from  the mean
vector  and  the covariance  matrix~\cite{Rasmussen:2005:GPM:1162254}.
For  instance,  let $\mathbf{x}=[\mathbf{x}_s,\mathbf{x}_{ns}]$  where
$\mathbf{x}_s$ and  $\mathbf{x}_{ns}$ are  the selected  variables and
the  non-selected  variables  respectively,  the mean  vector  can  be
written as
\begin{eqnarray}\label{margi:mean}
  \hat{\boldsymbol{\mu}}=[\boldsymbol{\mu}_s,\boldsymbol{\mu}_{ns}]^T
\end{eqnarray}
and the covariance matrix as
\begin{eqnarray}\label{margi:cov}
  \boldsymbol{\Sigma}=\left[\begin{array}{cc}
      \boldsymbol{\Sigma}_{s,s} & \boldsymbol{\Sigma}_{s,ns}\\
      \boldsymbol{\Sigma}_{ns,s} & \boldsymbol{\Sigma}_{ns,ns}
      \end{array}\right].
\end{eqnarray}
The  marginalization  over  the   non-selected  variables  shows  that
$\mathbf{x}_s$  is  also  a  Gaussian distribution  with  mean  vector
$\boldsymbol{\mu}_s$          and           covariance          matrix
$\boldsymbol{\Sigma}_{s,s}$.  Hence,  once the full model  is learned,
all the sub-models  built with a subset of the  original variables are
available at no computational cost.

\begin{algorithm}[t]\footnotesize
  \caption{NPFS pseudo code}
  \label{algo:npfs}
  \begin{algorithmic}[1]
    \REQUIRE $\mathcal{S}$, $k$, \texttt{delta}, \texttt{maxvariable}
    \STATE \rev{Randomly cut $\mathcal{S}$ into $k$ subsets such as $\mathcal{S}_1\cup\ldots\cup\mathcal{S}_k=\mathcal{S}$ and $\mathcal{S}_i\cap\mathcal{S}_j=\emptyset$}
    \STATE Learn the full GMM with $\mathcal{S}$
    \STATE Initialize the set of selected  variables $\varphi_s$ to empty set ($\left|\varphi_s\right|=0$) and available variables  $\varphi_a$ to the original set of variables ($\left|\varphi_a\right|=d$) 
    \WHILE{$\left|\varphi_s\right| \leq$ \texttt{maxvariable}}
    \FORALL{$\mathcal{S}_u\subset\mathcal{S}$}
    \STATE Update the model using eq.~(\ref{update:prop}), (\ref{update:mean}) and (\ref{update:cov}) \rev{(or their loocv counterparts)} according to $\mathcal{S}_u$
    \FORALL{$s \subset \varphi_a$}
    \STATE Compute the classification rate on $\mathcal{S}_u$ for each set of variables $\varphi_s \cap s$ using the marginalization properties
    \ENDFOR
    \ENDFOR
    \STATE Average the  classification rate over the $k$-fold
    \IF{Improvement in terms of  classification rate w.r.t. previous iteration is lower than \texttt{delta}}
    \BREAK
    \ELSE
    \STATE Add the variable $s$ corresponding to the maximum  classification rate to $\varphi_s$ and remove it from $\varphi_a$
    \ENDIF
    \ENDWHILE
  \end{algorithmic}
\end{algorithm}

\section{Experimental results}
\label{sec:exp}
\subsection{Data}

Two data sets have been used  in the experiments. The first data set
has  been acquired  in the  region  surrounding the  volcano Hekla  in
Iceland by the AVIRIS sensor.  157 spectral channels from 400 to 1,840
nm were recorded.  12 classes have  been defined for a total of 10,227
referenced pixels.  The second data set has been acquired by the ROSIS
sensor  during  a  flight  campaign over  Pavia,  nothern  Italy.  103
spectral channels  were recorded from  430 to  860 nm. 9  classes have
been defined for a total of 42776 referenced pixels.

For each  data set,  50, 100  and 200 training  pixels per  class were
randomly selected  and the remaining  referenced pixels were  used for
the validation. 50 repetitions were  done for which a new training set
have  been generated \rev{randomly}.  

\subsection{Competitive methods}
Several  conventional  feature selection  methods  have  been used  as
baseline.
\begin{itemize}
\item    Recursive   Feature    Elimination   (RFE)    for   nonlinear
  SVM~\cite{tuia2009classification}.   In the  experiment, a  Gaussian
  kernel was used.
\item  Linear SVM  with  $\ell_1$ (SVM$_{\ell_1}$)  constraint on  the
  feature  vector~\cite{tuia2014automatic}  based   on  the  LIBLINEAR
  implementation~\cite{REF08a}.
\item  To  overcome  the  limitation  of  the  linear  model  used  in
  LIBLINEAR,  an explicit  computation of  order 2  polynomial feature
  space     has     been      used     together     with     LIBLINEAR
  (SVM$_{\ell_1}^p$). Formally, a  nonlinear transformation $\phi$ has
  been apply on the original samples:
  \begin{eqnarray*}
    \mathbb{R}^d &  \to & \mathbb{R}^{p}\\
    \mathbf{x}=\left[x_1,\ldots,x_d\right] & \mapsto & \phi(\mathbf{x})=\left[x_1,\ldots,x_d,x_1^2,x_1x_2,\ldots,x_1x_d,x_2^2,x_2x_3,\ldots,x_d^2\right]
  \end{eqnarray*}
  with  $p=\binom{2+d}{2}$.  For  Hekla data  and University  of Pavia
  data, the  dimension $p$ of the  projected space is 12561  and 5460,
  respectively.
\end{itemize}

For comparison, a SVM with a  Gaussian kernel and a order 2 polynomial
kernel  classifier, based  on  the LIBSVM~\cite{CC01a},  with all  the
variables have been used too.

For the  linear/nonlinear SVM,  the penalty  parameter and  the kernel
hyperparameters were selected using 5-fold cross-validation. For NPFS,
the threshold \rev{(\texttt{delta}  in Algorithm~\ref{algo:npfs})} was
set to 0.5\%  and the maximum number of extracted  features was set to
20. The estimation of the error has been computed with a leave-one-out
CV  ($n$-NPFS) and  a  5-fold  CV (5-NPFS).   Each  variable has  been
standardized  before   the  processing  (i.e.,  zero   mean  and  unit
variance).

\begin{table*}\footnotesize
  \centering
  \caption{Classification accuracies for Hekla data set. The results correspond to the mean value and variance of the overall accuracy over the 50 repetitions. The best result for each training setup is reported in bold face. $n$-NPFS and 5-NPFS correspond to the NPFS computed with the leave-one-out and 5-fold  cross-validation, respectively. RFE, SVM$_{\ell_1}$ and  SVM$_{\ell_1}^p$ correspond to the recursive feature extraction SVM, the linear SVM with $\ell_1$ constraint and the linear SVM with $\ell_1$ with the explicit order 2 polynomial feature space, respectively. SVM$_{\text{poly}}$ and  SVM$_\text{gauss}$ correspond to the conventional nonlinear SVM with a order 2 polynomial kernel  and Gaussian kernel, respectively.}
  \label{tab:hekla}
  \begin{tabular}{c|ccccccc}
    \toprule
    $n_c$ &$n$-NPFS &5-NPFS & RFE & SVM$_{\ell_1}$ & SVM$_{\ell_1}^p$ & SVM$_{\text{poly}}$& SVM$_\text{gauss}$\\
    \midrule
    50 & \bf 92.5 $\pm$ 1.2 &92.4 $\pm$ 1.2  & 90.2 $\pm$ 1.8& 90.3 $\pm$ 1.0 & 91.6 $\pm$ 0.6 & 84.6 $\pm$ 1.6 & 90.4 $\pm$ 1.6\\
    \midrule
    100 & 94.8 $\pm$ 0.7 &94.6 $\pm$ 0.6 &\bf 95.6 $\pm$ 0.3 & 93.9 $\pm$ 0.5 & 94.8 $\pm$ 0.1 &91.4 $\pm$ 0.4 &\bf 95.6 $\pm$ 0.3 \\
    \midrule
    200 & 95.9 $\pm$ 0.3 & 95.8 $\pm$ 0.3  & \bf 96.8 $\pm$ 1.1  & 95.6 $\pm$ 0.1& 96.3 $\pm$ 0.1  &95.5 $\pm$ 0.1 & \bf 96.8 $\pm$ 1.1 \\
    \bottomrule
    \end{tabular}
\end{table*}

\begin{table*}\footnotesize
  \centering
  \caption{Classification accuracies for University of Pavia data set. Same notations than in Table~\ref{tab:hekla}.}
  \label{tab:uni}
  \begin{tabular}{c|ccccccc}
     \toprule
    $n_c$ &$n$-NPFS &5-NPFS & RFE & SVM$_{\ell_1}$ & SVM$_{\ell_1}^p$ & SVM$_{\text{poly}}$& SVM$_\text{gauss}$\\
    \midrule
    50 & 82.2 $\pm$ 4.4 & 83.4 $\pm$ 7.6 & 84.7 $\pm$ 4.0 &75.1 $\pm$ 2.5 &81.0 $\pm$ 2.8 & 82.9 $\pm$ 3.4 & \bf 84.8 $\pm$ 3.4\\
    \midrule
    100 & 86.3 $\pm$ 3.2 & 85.9 $\pm$ 3.1  & \bf 88.4 $\pm$ 0.9 &77.3 $\pm$ 1.4 &83.6 $\pm$ 1.3 & 86.5 $\pm$ 1.6 & 88.4 $\pm$ 1.4\\
    \midrule
    200 & 87.7 $\pm$ 3.1 & 87.9 $\pm$ 1.9 & \bf 90.8 $\pm$ 0.3 &78.5 $\pm$ 0.7 & 85.5 $\pm$ 0.4 & 88.8 $\pm$ 0.6 & \bf 90.8 $\pm$ 0.3\\
    \bottomrule
  \end{tabular}
\end{table*}

\begin{figure}
  \centering
  \begin{tikzpicture}
    \begin{axis}[ylabel=$\bar{n}_s$,
      symbolic x coords={0,50,100,200,250},
      enlargelimits=0.15,legend pos=outer north east,xtick=data,legend cell align=left,
      ybar,
      bar width=5pt,grid,xlabel=Number of training samples per class,small,area legend]
      \addplot coordinates {(50,7.6) (100,8.5) (200,8.8)};
      \addplot coordinates {(50,7.3) (100,8.2) (200,8.6)};
      \addplot coordinates {(50,157) (100,157) (200,157)};
      \addplot coordinates {(50,154.7) (100,155.2) (200,155.3)};
      \addplot[red,very thick,sharp plot,update limits=false]coordinates {(0,157) (250,157)};
      \legend{{\footnotesize $n$-NPFS},{\footnotesize 5-NPFS},{\footnotesize SVM$_{\ell_1}$},{\footnotesize RFE}}
    \end{axis}
  \end{tikzpicture}
  \caption{Mean number $\bar{n}_s$ of selected features  for the different methods for Hekla data set. The red line indicates the original number of spectral features. Projected $\ell_1$ SVM is not reported because the mean number of extracted features was too high (e.g., 6531 for $n_c$=50).}
  \label{fig:meanS:hekla}
\end{figure}
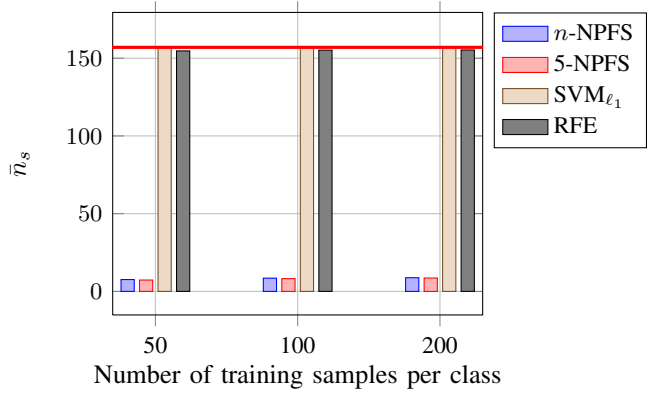

\begin{figure}
  \centering
  \begin{tikzpicture}
    \begin{axis}[ylabel=$\bar{n}_s$,
      symbolic x coords={0,50,100,200,250},
      enlargelimits=0.15,legend pos=outer north east,xtick=data,legend cell align=left,
      ybar,
      bar width=5pt,grid,xlabel=Number of training samples per class,small,area legend]
      \addplot coordinates {(50,6.4) (100,6.5) (200,6.9)};
      \addplot coordinates {(50,6.1) (100,6.6) (200,7.0)};
      \addplot coordinates {(50,61.6) (100,78.3) (200,92.3)};
      \addplot coordinates {(50,98.4) (100,99.1) (200,99.3)};      
      \addplot[red,very thick,sharp plot,update limits=false]coordinates {(0,103) (250,103)};
      \legend{{\footnotesize $n$-NPFS},{\footnotesize 5-NPFS},{\footnotesize SVM$_{\ell_1}$},{\footnotesize RFE}}
    \end{axis}
  \end{tikzpicture}
  \caption{Mean number $\bar{n}_s$ of selected features  for the different methods for University of Pavia data set. The red line indicates the original number of spectral features. Projected $\ell_1$ SVM is not reported because the mean number of extracted features was too high (e.g., 5110 for $n_c$=50).}
  \label{fig:meanS:uni}
\end{figure}
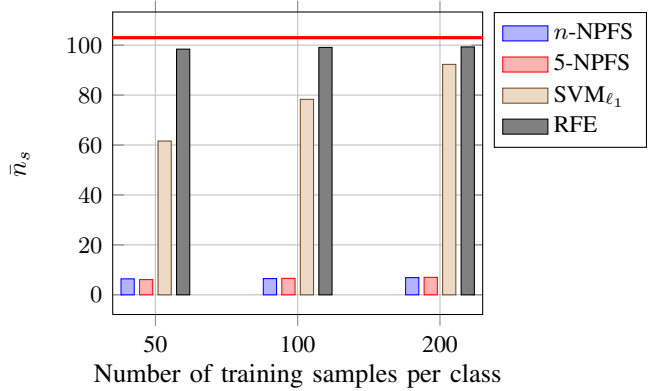

\subsection{Results}
The mean accuracies and the variance  over the 50 runs are reported in
Table~\ref{tab:hekla}  and Table~\ref{tab:uni}.   The mean  numbers of
extracted  features   for  the  different  methods   are  reported  in
Figure~\ref{fig:meanS:hekla} and Figure~\ref{fig:meanS:uni}.

From the  tables, it can  be seen that there  is no difference  in the
results obtained  with $n$-NPFS or  $5$-NPFS. They perform  equally on
both  data sets  in  terms  of classification  accuracy  or number  of
extracted  features.  However,  $5$-NPFS is  much faster  in terms  of
computation time.

RFE  and  SVM$_\text{gauss}$ provide  the  best  results in  terms  of
classification   accuracy,  except   for  the   Hekla  data   set  and
$n_c=50$. From the figure, it can be seen that the number of extracted
features  is almost  equal to  original number  of spectral  features,
meaning   that   in   these   experiments   RFE   is   equivalent   to
SVM$_\text{gauss}$.   Hence, RFE  was  not able  to extract  \rev{few}
relevant spectral features.

$\ell_1$  SVM  applied  on  the original  features  or  the  projected
features  is not  able  to  extract relevant  features.   In terms  of
classification accuracy, the linear SVM  does not perform well for the
University of  Pavia data set.   Nonlinear $\ell_1$ SVM  provides much
better results  for both data sets.   In comparison to the  non sparse
nonlinear SVM  computed with  an order  2 polynomial  kernel, $\ell_1$
nonlinear SVM performs better in  terms of classification accuracy for
the Hekla  data while it  performs worst  for the University  of Pavia
data.

In  terms of  number of  extracted  features, NPFS  provides the  best
results, by far, with an average number of extracted features equal to
5\% of  the original number. All  the others methods were  not able to
extract  few  features  without  decreasing  drastically  the  overall
accuracy.  For instance,  for the Hekla data set and  $n_c=50$, only 7
spectral features  are used  to build  the GMM and  leads to  the best
classification  accuracy. A  discussion on  the extracted  features is
given in the next subsection.

\begin{figure}
  \centering
  \includegraphics[width=0.4\linewidth]{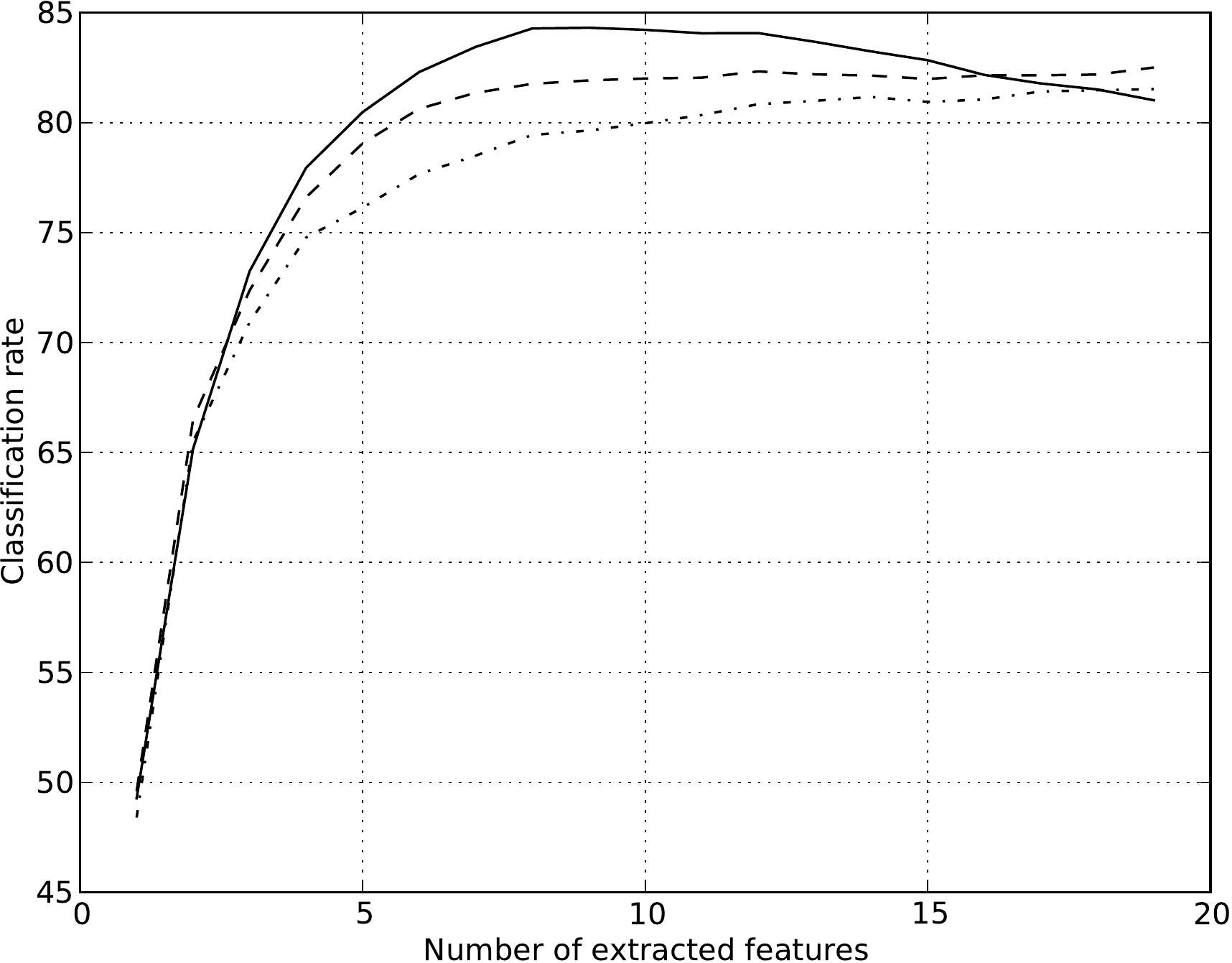}
  \caption{\rev{Classification rate in function of the number of extracted features. Continuous line corresponds to 5-NPFS, dashed line to SVM with a Gaussian kernel and dash-doted line to a linear SVM.}}
  \label{fig:oa_nf}
\end{figure}

\rev{The figure~\ref{fig:oa_nf}  presents the  averaged classification
  rate of 5-NPFS, SVM with a  Gaussian kernel and a linear SVM applied
  on the selected features with 5-NPFS.  20 repetitions have been done
  on the University data set with $n_c$=50. The optimal parameters for
  SVM   and    linear   SVM   have   been    selected   using   5-fold
  cross-validation. From  the figure,  it can be  seen that  the three
  algorithms  have similar  trends.  When  the number  of features  is
  relatively low (here lower than 15)  GMM performs the best, but when
  the  number of  features increases  too  much, SVM  (non linear  and
  linear) performs better in terms  of classification accuracy.  It is
  worth  noting   that  such   observations  are  coherent   with  the
  literature: SVM are known to perform well in high dimensional space,
  while GMM is  more affected by the dimension.  Yet,  NPFS is able to
  select  relevant  spectral  variables,  for  itself,  or  for  other
  (possibly) stronger algorithms.

}

\rev{The mean processing time for the University of Pavia data set for
  several training  set sizes  is reported in  Table~\ref{tab:pt}.  It
  includes   parameter   optimization   for   SVM$_\text{gauss}$   and
  SVM$_{\ell_1}$.    Note   that   the   RFE   consists   in   several
  SVM$_\text{gauss}$  optimization,  one   for  each  feature  removed
  (hence,  if 3  features are  removed,  the mean  processing time  is
  approximately multiply by 3). It can  be seen that the 5-NPFS method
  is a  little influenced  by the  size of the  training set:  what is
  important is the number of (extracted) variables.  For $n_s=50$, the
  processing  time  is slightly  higher  because  of overload  due  to
  parallelization procedure.  $n$-NPFS is  the more demanding in terms
  of processing time  and thus should be used only  when the number of
  training  samples is  very  limited.  Finally,  it  is important  to
  underline that  the NPFS is implemented  in Python while SVM  used a
  state of the art implementation in C++~\cite{CC01a}.}

\begin{table}
  \centering
  \caption{\rev{Mean processing time in second in function of the number of samples per class for the University of Pavia data set. 20 repetitions have been done on  laptop with 8Gb of RAM and
      Intel(R) Core(TM)  i7-3667U CPU  @ 2.00GHz processor.}}
  \label{tab:pt}
  \begin{tabular}{ccccc}
    \toprule
    $n_s$ & 50 & 100 & 200 & 400\\
    \midrule
    SVM$_\text{gauss}$ & 11 & 40 & 140 & 505\\
    SVM$_{\ell_1}$ & 52 & 115 & 234 & 498\\
    $n$-NPFS & 242 & 310 & 472 & 883\\
    5-NPFS & 35  & 31 & 29 & 43\\
    \bottomrule
  \end{tabular}
\end{table}

From these experiments, and from a practical viewpoint, NPFS is a good
compromise between high classification accuracy and sparse modeling. 

\subsection{Discussion}
The extracted  channels by 5-NPFS  and $n$-NPFS were compared  for one
training set  of the University of  Pavia data set: two  channels were
the same  for both methods,  780nm and  776nm; two channels  were very
close, 555nm  and 847nm for 5-NPFS  and 551nm and 855nm  for $n$-NPFS;
one channel was  close, 521nm for 5-NPFS and 501nm  for $n$-NPFS.  The
other  channel selected  by  $n$-NPFS  is 772nm.   If  the process  is
repeated, the  result is  terms of selected  features by  $n$-NPFS and
5-NPFS  is similar:  on  average  35\% of  the  selected features  are
identical (not  necessarily the  first ones)  and the  others selected
features are close in terms of wavelength.

\begin{figure}
  \centering
  \includegraphics[width=0.45\linewidth]{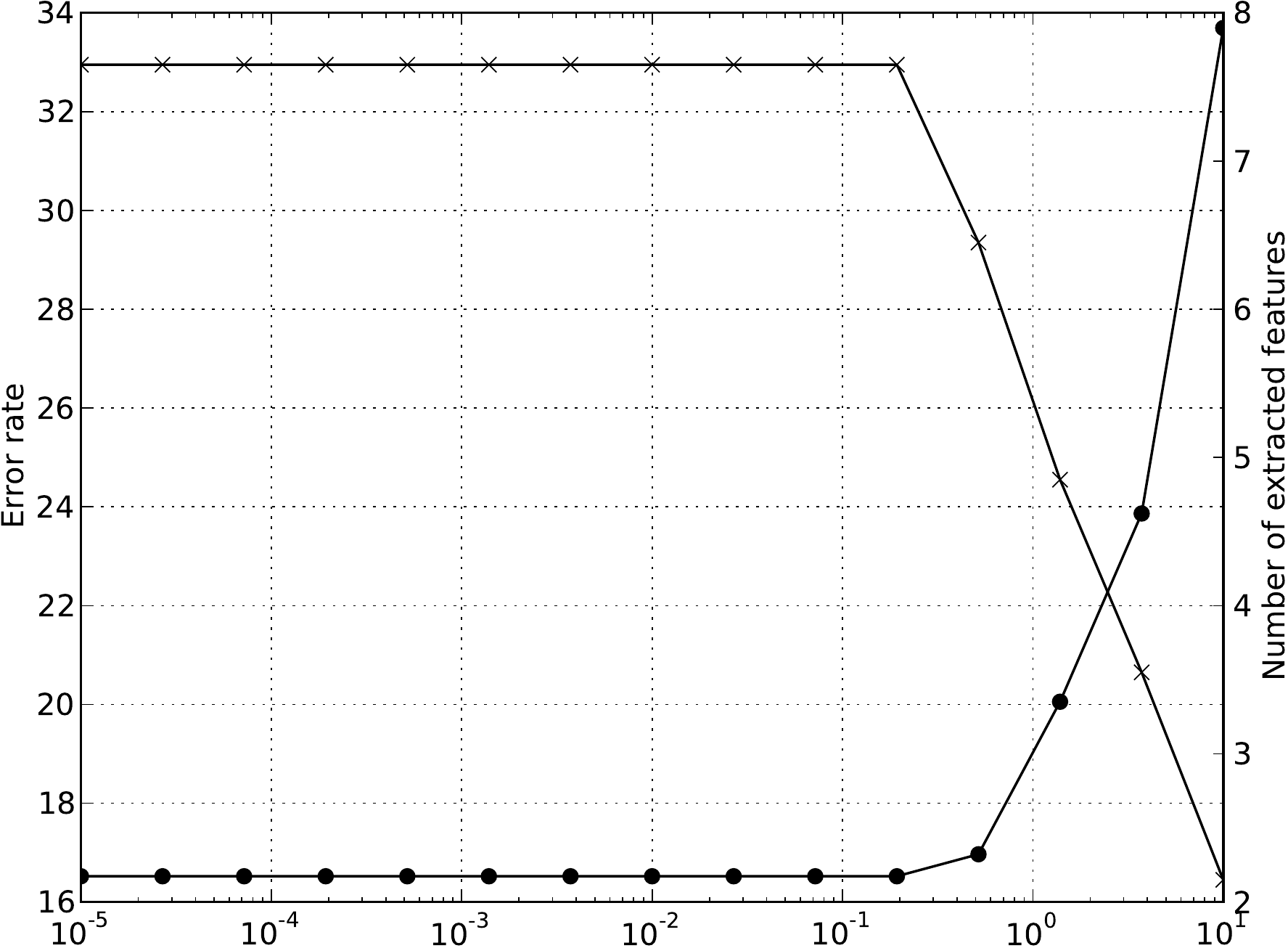}
  \caption{\rev{The  dotted line  and the  crossed line  represent the
      mean  error  rate and  the  mean  number of  selected  features,
      respectively, as a function of delta. The simulation was done on
      the  University of  Pavia data  set, with  $n_c=50$ and  for the
      5-NPFS algorithm.}}
  \label{fig:oa_delta}
\end{figure}

\rev{The   influence  of   the  parameter   \texttt{delta}  has   been
  investigated on the University of Pavia data set. 20 repetitions have
  been  done  with  $n_c=50$  for several  values  of  \texttt{delta}.
  Results are reported on  figure~\ref{fig:oa_delta}. From the figure,
  it  can be  seen that  when  delta is  set  to a  value larger  than
  approximately 1\%,  the algorithm stops  too early and the  number of
  selected features  is too low  to perform well.   Conversely, setting
  \texttt{delta} to a  small value does not  change the classification
  rate, a plateau  being reached for \texttt{delta}  lower than 0.5\%.
  In  fact,  because of  the  ``Hughes  phenomenon'', adding  spectral
  features  to  the  GMM  will  first  lead  to  an  increase  of  the
  classification  rate   but  then  (after  a   possible  plateau)  the
  classification rate will decrease,  i.e., the improvement after two
  iterations of the algorithm will be negative.}

\begin{figure}
  \centering
  \includegraphics[width=0.45\textwidth]{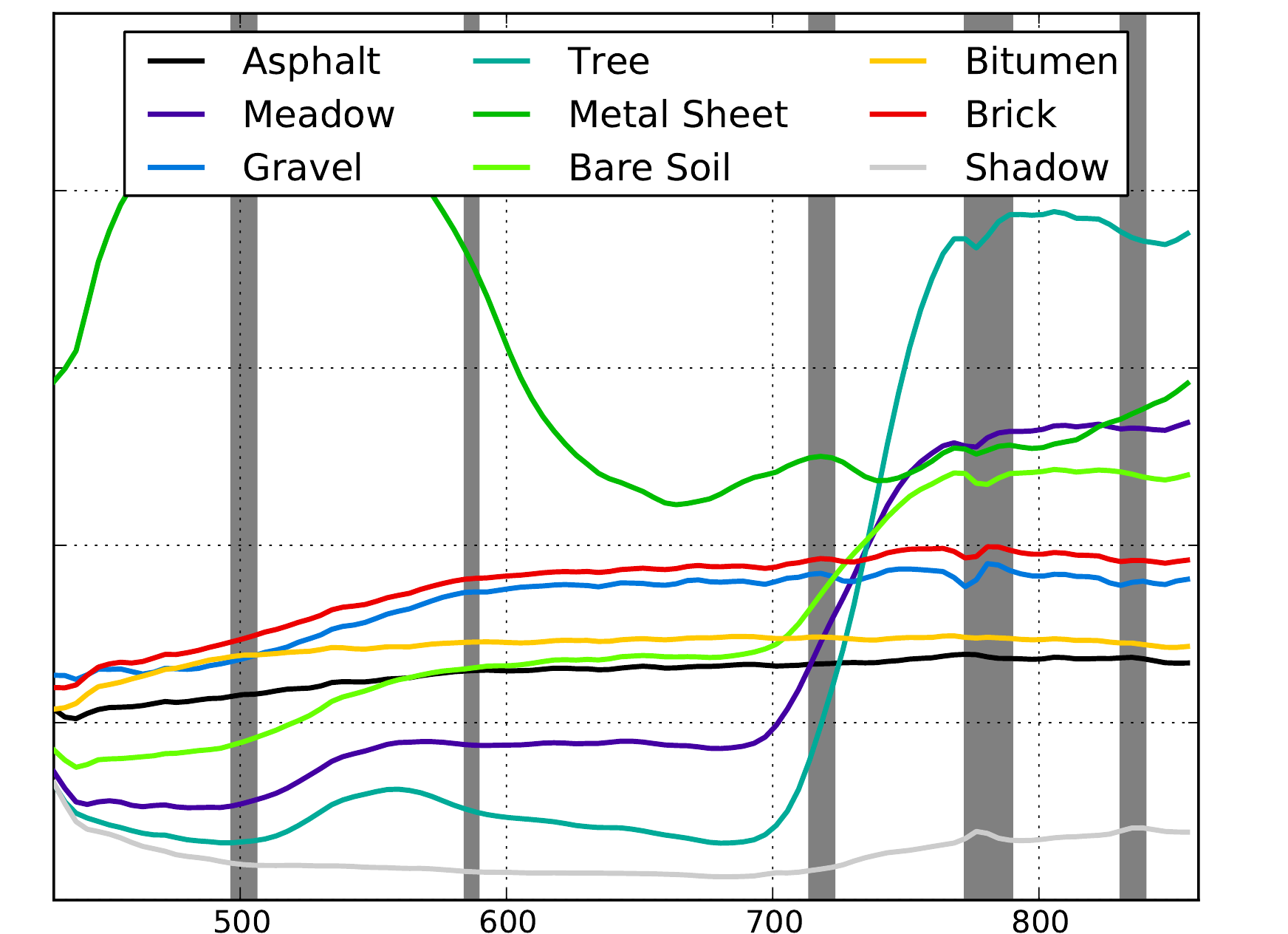}
  \caption{Most selected  spectral domain for the  University of Pavia
    data set.  Gray bars correspond to  the most selected parts of the
    spectral domain.   Horizontal axis  corresponds to  the wavelength
    (in nanometers).  The  mean value of each class  is represented in
    continuous colored lines.}
  \label{fig:select}
\end{figure}

Figure~\ref{fig:select} presents  the most  selected features  for the
University of Pavia  data set. 1000 random repetitions  have been done
with  $n_c$=200  and the  features  shaded  in  the figure  have  been
selected  at  least 10\%  times  (i.e.,  100  times over  1000)  using
5-NPFS. Five spectral domains can  be identified, two from the visible
range and three  from the near infrared range.  In  particular, it can
be seen  that spectral channels  from the red-edge part  are selected.
The  width of  the spectral  domain indicates  the variability  of the
selection.  The high correlation between adjacent spectral bands makes
the variable selection  ``unstable'', e.g., for a  given training set,
the channel  $t$ would be  selected but for another  randomly selected
training set it might be the channel  $t+1$ or $t-1$.  It is clearly a
limitation of the proposed approach.

To conclude  this discussion, similar spectral  channels are extracted
with   $n$-NPFS  and   5-NPFS,  while   the  latter   is  much   times
faster. \rev{Hence,  $n$-NPFS should  be only  used when  very limited
  number of samples is available}.   A certain variability is observed
in the selection of the spectral  channels due to the high correlation
of adjacent spectral channels and the step-wise nature of the method.

\section{Conclusion}
\label{sec:conclusion}
A  nonlinear   parsimonious  feature   selection  algorithm   for  the
classification of  hyperspectral images and the  selection of spectral
variables  has  been  presented.    Using  a  Gaussian  mixture  model
classifier, spectral variables are  extracted iteratively based on the
cross-validation estimate  of the  classification rate.   An efficient
implementation is proposed that takes  into account some properties of
Gaussian mixture  model: a fast update  of the model parameters  and a
fast access  to the  sub-models.  Experimental  results show  that the
proposed  method  is  able  to   select  few  relevant  features,  and
outperform standard SVM-based sparse algorithms while reaching similar
classification rates to those  obtained with SVM. \rev{Furthermore, in
  comparison  to SVM  based  feature  selection algorithm,  multiclass
  problem  is handled  by the  GMM  making the  interpretation of  the
  extracted channels easier.}

More investigation are  needed to fully understand  which features are
extracted, since the method is purely statistical. If the red-edge has
been  identified,  the  others  extracted  features  are  not  clearly
interpretable. Moreover,  small variability  has been observed  due to
the  high  correlation  between   adjacent  bands  and  the  step-wise
procedure.   To  overcome  this   limitation,  a  continuous  interval
selection     strategy,     as      in~\cite{4069122},     will     be
investigated. \rev{Also,  a steepest-ascent  search strategy  could be
  used to make the final solution more stable}.

\rev{The  python  code  of  the  algorithm  is  available  freely  for
  download: \url{https://github.com/mfauvel/FFFS}.}

\section{Acknowledgment}
\rev{The authors would like to thank Professor P. Gamba, University of
  Pavia, for providing the University  of Pavia data set and Professor
  J.A.  Benediktsson,  University of Iceland, for  providing the Hekla
  data set.   They would like  also to  thank the reviewers  for their
  many helpful comments.}

\bibliographystyle{IEEEbib}
\bibliography{IEEEabrv,refs}

\begin{thebibliography}{10}

\bibitem{661089}
L.~O. Jimenez and D.~A. Landgrebe,
\newblock ``Supervised classification in high-dimensional space: geometrical,
  statistical, and asymptotical properties of multivariate data,''
\newblock {\em Systems, Man, and Cybernetics, Part C: Applications and Reviews,
  IEEE Transactions on}, vol. 28, no. 1, pp. 39 --54, feb 1998.

\bibitem{donoho}
D.~L. Donoho,
\newblock ``High-dimensional data analysis: the curses and blessing of
  dimensionality,''
\newblock in {\em AMS Mathematical challenges of the 21st century}, 2000.

\bibitem{hughes}
G.~F. Hughes,
\newblock ``On the mean accuracy of statistical pattern recognizers,''
\newblock {\em {IEEE} Trans. Inf. Theory}, vol. IT-14, pp. 55--63, January
  1968.

\bibitem{fauvel:hal-00737075}
M.~Fauvel, Y.~Tarabalka, J.~A. Benediktsson, J.~Chanussot, and J.~Tilton,
\newblock ``{Advances in Spectral-Spatial Classification of Hyperspectral
  Images},''
\newblock {\em Proceedings of the IEEE}, vol. 101, no. 3, pp. 652--675, Mar.
  2013.

\bibitem{DR:guided:tour}
C.~J.~C. Burges,
\newblock ``Dimension reduction: A guided tour,''
\newblock {\em Foundations and Trends in Machine Learning}, vol. 2, no. 4, pp.
  275--365, 2010.

\bibitem{kernel:methods:rs}
G.~Camps-Valls and L.~Bruzzone, Eds.,
\newblock {\em Kernel Methods for Remote Sensing Data Analysis},
\newblock Wiley, 2009.

\bibitem{1294808}
A.~Cheriyadat and L.~M. Bruce,
\newblock ``Why principal component analysis is not an appropriate feature
  extraction method for hyperspectral data,''
\newblock in {\em Geoscience and Remote Sensing Symposium, 2003. Proceedings},
  July 2003, vol.~6, pp. 3420--3422 vol.6.

\bibitem{5942156}
A.~Villa, J.~A. Benediktsson, J.~Chanussot, and C.~Jutten,
\newblock ``Hyperspectral image classification with independent component
  discriminant analysis,''
\newblock {\em {IEEE} Trans. Geosci. Remote Sens.}, vol. 49, no. 12, pp.
  4865--4876, Dec 2011.

\bibitem{fauvel:hal-00283769}
M.~Fauvel, J.~A. Benediktsson, J.~Chanussot, and J.~R. Sveinsson,
\newblock ``{Spectral and Spatial Classification of Hyperspectral Data Using
  SVMs and Morphological Profiles},''
\newblock {\em {IEEE} Trans. Geosci. Remote Sens.}, vol. 46, no. 11 - part 2,
  pp. 3804--3814, Oct. 2008.

\bibitem{6851112}
F.E. Fassnacht, C.~Neumann, M.~Forster, H.~Buddenbaum, A.~Ghosh, A.~Clasen,
  P.K. Joshi, and B.~Koch,
\newblock ``Comparison of feature reduction algorithms for classifying tree
  species with hyperspectral data on three central european test sites,''
\newblock {\em Selected Topics in Applied Earth Observations and Remote
  Sensing, IEEE Journal of}, vol. 7, no. 6, pp. 2547--2561, June 2014.

\bibitem{6947182}
M.~Lothode, V.~Carrere, and R.~Marion,
\newblock ``Identifying industrial processes through vnir-swir reflectance
  spectroscopy of their waste materials,''
\newblock in {\em Geoscience and Remote Sensing Symposium (IGARSS), 2014 IEEE
  International}, July 2014, pp. 3288--3291.

\bibitem{4069122}
S.B. Serpico and G.~Moser,
\newblock ``Extraction of spectral channels from hyperspectral images for
  classification purposes,''
\newblock {\em {IEEE} Trans. Geosci. Remote Sens.}, vol. 45, no. 2, pp.
  484--495, Feb 2007.

\bibitem{chein2007hyperspectral}
C.-I Chang,
\newblock {\em Hyperspectral Data Exploitation: Theory and Applications},
\newblock Wiley, 2007.

\bibitem{934069}
S.B. Serpico and L.~Bruzzone,
\newblock ``A new search algorithm for feature selection in hyperspectral
  remote sensing images,''
\newblock {\em Geoscience and Remote Sensing, IEEE Transactions on}, vol. 39,
  no. 7, pp. 1360--1367, Jul 2001.

\bibitem{5161332}
L.~Bruzzone and C.~Persello,
\newblock ``A novel approach to the selection of spatially invariant features
  for the classification of hyperspectral images with improved generalization
  capability,''
\newblock {\em Geoscience and Remote Sensing, IEEE Transactions on}, vol. 47,
  no. 9, pp. 3180--3191, Sept 2009.

\bibitem{4317535}
A.C. Jensen and R.~Solberg,
\newblock ``Fast hyperspectral feature reduction using piecewise constant
  function approximations,''
\newblock {\em Geoscience and Remote Sensing Letters, IEEE}, vol. 4, no. 4, pp.
  547--551, Oct 2007.

\bibitem{lebris:fs}
A.~Le Bris, N.~Chehata, X.~Briottet, and N.~Paparoditis,
\newblock ``Use intermediate results of wrapper band selection methods : a
  first stpe toward the optimisation of spectral configuration for land cover
  classifications,''
\newblock in {\em Proc. of the IEEE WHISPERS'14}, 2014.

\bibitem{tuia2009classification}
D.~Tuia, F.~Pacifici, M.~Kanevski, and W.~J. Emery,
\newblock ``Classification of very high spatial resolution imagery using
  mathematical morphology and support vector machines,''
\newblock {\em Geoscience and Remote Sensing, IEEE Transactions on}, vol. 47,
  no. 11, pp. 3866--3879, 2009.

\bibitem{tuia2014automatic}
D.~Tuia, M.~Volpi, M.~Dalla~Mura, A.~Rakotomamonjy, and R.~Flamary,
\newblock ``Automatic feature learning for spatio-spectral image classification
  with sparse {SVM},''
\newblock {\em {IEEE} Trans. Geosci. Remote Sens.}, vol. PP, no. 99, pp. 1--13,
  2014.

\bibitem{5440922}
G.~Camps-Valls, J.~Mooij, and Bernhard Scholkopf,
\newblock ``Remote sensing feature selection by kernel dependence measures,''
\newblock {\em Geoscience and Remote Sensing Letters, IEEE}, vol. 7, no. 3, pp.
  587--591, July 2010.

\bibitem{Ferraty01122010}
F.~Ferraty, P.~Hall, and P.~Vieu,
\newblock ``Most-predictive design points for functional data predictors,''
\newblock {\em Biometrika}, vol. 97, no. 4, pp. 807--824, 2010.

\bibitem{hastie2001elements}
T.~Hastie, R.~Tibshirani, and J.~H. Friedman,
\newblock {\em The Elements of Statistical Learning: Data Mining, Inference,
  and Prediction},
\newblock Springer series in statistics. Springer, 2001.

\bibitem{Rasmussen:2005:GPM:1162254}
C.~E. Rasmussen and C.~K.~I. Williams,
\newblock {\em Gaussian Processes for Machine Learning (Adaptive Computation
  and Machine Learning)},
\newblock The MIT Press, 2005.

\bibitem{REF08a}
R.-E. Fan, K.-W. Chang, C.-J. Hsieh, X.-R. Wang, and C.-J. Lin,
\newblock ``{LIBLINEAR}: A library for large linear classification,''
\newblock {\em Journal of Machine Learning Research}, vol. 9, pp. 1871--1874,
  2008.

\bibitem{CC01a}
Chih-C. Chang and C.-J. Lin,
\newblock ``{LIBSVM}: A library for support vector machines,''
\newblock {\em ACM Transactions on Intelligent Systems and Technology}, vol. 2,
  pp. 27:1--27:27, 2011,
\newblock Software available at \url{http://www.csie.ntu.edu.tw/~cjlin/libsvm}.

\end{thebibliography}
\end{document}